\documentclass{article}

\usepackage{algorithmic}
\usepackage{fancybox}

\usepackage[preprint]{neurips_2026}

\usepackage[utf8]{inputenc} %
\usepackage[T1]{fontenc}    %
\usepackage{hyperref}       %
\usepackage{url}            %
\usepackage{booktabs}       %
\usepackage{amsfonts}       %
\usepackage{nicefrac}       %
\usepackage{microtype}      %
\usepackage{xcolor}         %
\usepackage{todonotes}
\usepackage[most]{tcolorbox}

\usepackage{amsmath}
\usepackage{amssymb}
\usepackage{bm}
\usepackage{mathtools}

\def\vtheta{{\bm{\theta}}}

\def\vd{{\bm{d}}}

\def\vg{{\bm{g}}}

\def\vlambda{{\bm{\lambda}}}

\def\mA{{\bm{A}}}

\def\mD{{\bm{D}}}

\def\mG{{\bm{G}}}

\def\mI{{\bm{I}}}

\def\mO{{\bm{O}}}
\def\mP{{\bm{P}}}
\def\mQ{{\bm{Q}}}
\def\mR{{\bm{R}}}
\def\mS{{\bm{S}}}

\def\mZ{{\bm{Z}}}

\def\mTheta{{\bm{\Theta}}}

\DeclareMathAlphabet{\mathsfit}{\encodingdefault}{\sfdefault}{m}{sl}
\SetMathAlphabet{\mathsfit}{bold}{\encodingdefault}{\sfdefault}{bx}{n}

\def\sI{{\mathbb{I}}}

\def\sR{{\mathbb{R}}}

\DeclareSymbolFont{bbold}{U}{bbold}{m}{n}
\DeclareSymbolFontAlphabet{\mathbbold}{bbold}

\DeclareMathOperator{\diag}{diag}

\DeclareMathOperator{\Mat}{Mat}

\DeclarePairedDelimiterX{\KLdivx}[2]{(}{)}{%
  #1\;\delimsize\|\;#2%
}

\usepackage{nicefrac}       %
\usepackage{microtype}      %
\usepackage[dvipsnames,table]{xcolor}
\definecolor{TUblue}{RGB}{0,105,170}
\usepackage{booktabs}
\usepackage{doi}
\usepackage{cleveref}

\usepackage{comment}
\newcommand{\wu}[1]{{\color{red} [\textbf{Wu:} #1]}}

\newcommand{\felix}[1]{{\color{green!50!black} [\textbf{Felix:} #1]}}

{}
{}
{}
{}
{}

\DeclareMathOperator{\Diag}{Diag}
\DeclareMathOperator{\qr}{qr}

\usepackage{pifont}         %

\definecolor{vlightgray}{gray}{0.87}

\definecolor{tblue}{HTML}{1F77B4}
\definecolor{torange}{HTML}{FF7F0E}
\definecolor{tgreen}{HTML}{2CA02C}
\definecolor{tred}{HTML}{FF0000}

\definecolor{linkcolor}{HTML}{991408}  %
\definecolor{citecolor}{HTML}{2E7E2A}  %
\definecolor{filecolor}{HTML}{131877}  %
\definecolor{menucolor}{HTML}{727500}  %
\definecolor{runcolor} {HTML}{137776}  %
\definecolor{urlcolor} {HTML}{0a2bbf}  %
\hypersetup{colorlinks=true,linkcolor=urlcolor,citecolor=urlcolor,filecolor=urlcolor,menucolor=urlcolor,runcolor=urlcolor,urlcolor=urlcolor}

\newcommand{\highlight}[1]{\textcolor{red!75!black}{#1}}
\newcommand{\highlightii}[1]{\textcolor{blue!75!black}{#1}}

\newcommand{\runtimeCell}[3]{%
  \pgfmathsetmacro{\rcRatio}{#2/#3}%
  \pgfmathsetmacro{\rcAbsDev}{min(abs(\rcRatio-1), 0.2)}%
  \pgfmathtruncatemacro{\rcIntensity}{\rcAbsDev*500}%
  \ifdim\rcRatio pt<1pt
    \edef\rcAct{\noexpand\cellcolor{green!\rcIntensity!yellow}}%
  \else
    \edef\rcAct{\noexpand\cellcolor{red!\rcIntensity!yellow}}%
  \fi
  \rcAct #1%
}
\newcommand{\runtimeColorbar}{%
  \begin{tikzpicture}[xscale=1.1]
    \foreach \i in {0,1,...,79} {%
      \pgfmathsetmacro{\rcR}{0.8 + \i/200}%
      \pgfmathsetmacro{\rcD}{abs(\rcR-1)}%
      \pgfmathtruncatemacro{\rcI}{\rcD*500}%
      \ifdim\rcR pt<1pt
        \fill[green!\rcI!yellow] (\i*0.05, 0) rectangle ({(\i+1)*0.05}, 0.3);
      \else
        \fill[red!\rcI!yellow] (\i*0.05, 0) rectangle ({(\i+1)*0.05}, 0.3);
      \fi
    }%
    \draw (0,0) rectangle (4,0.3);
    \foreach \x in {1, 2, 3}
      \draw[line width=0.6pt] (\x, 0) -- (\x, 0.3);
    \foreach \v/\x in {0.8/0, 0.9/1, 1.0/2, 1.1/3, 1.2/4}
      \node[below, font=\small] at (\x, 0) {\v};
    \node[left, xshift=-0.3em] at (0, 0.15) {runtime ratio};
  \end{tikzpicture}%
}

\newcommand{\appref}[1]{\hyperref[#1]{Appendix~\ref*{#1}}}%

\def\Snospace~{\S{}}%
\newcommand{\hider}[1]{}

\title{
Scaling KL-Shampoo
}

\author{%
  Alan Milligan \\
  Mila \& Universit\'e de Montr\'eal \\
  \And
  Zikun Xu \\
  Microsoft \\
  \And
  Simon Lacoste-Julien \\
  Mila \& Universit\'e de Montr\'eal \\
  \AND
  Felix Dangel\thanks{Joint last authors.} \\
  Concordia University \& Mila \\
  \And
  Wu Lin\footnotemark[1]\, \thanks{Corresponding author: \href{mailto:yorker.lin@gmail.com}{yorker.lin@gmail.com}.}\\
  University of Central Florida \\
}

\usepackage{times}
\usepackage{hyperref}
\usepackage{url}
\usepackage{amsmath}
\usepackage{amssymb}
\usepackage{mathtools}
\usepackage{algorithm}
\usepackage{algorithmic}
\usepackage{natbib}
\usepackage{multirow, makecell}

\usepackage{caption}

\usepackage{mathtools} 
\usepackage{nicematrix}
\definecolor{hlblue}{RGB}{199,226,246}
\definecolor{hlpink}{RGB}{248,206,197}

\usepackage{tikz}
\usepackage[customcolors, norndcorners]{hf-tikz}

\hfsetfillcolor{white}
\hfsetbordercolor{red!75!black}

\usepackage{nicematrix}
\usetikzlibrary{decorations.pathreplacing}

\title{Reparametrizing Shampoo and SOAP for Subspace Basis Updates and BFloat16 Storage}

\begin{document}

\maketitle

\begin{abstract}
Shampoo-based methods, such as KL-Shampoo and SOAP, have demonstrated strong performance in training neural networks and rely on QR decomposition. Because existing QR implementations require single-precision (FP32) arithmetic and remain computationally expensive, these methods become time- and memory-intensive when their preconditioning matrices are large. Moreover, using BFloat16 (BFP16) storage to reduce memory usage can degrade the performance of Shampoo-based methods. We propose a reparametrization of the preconditioner that supports BFP16 storage and forms a complete basis by combining updated basis vectors with unchanged ones. By updating only part of the basis through QR decomposition in a subspace, our approach reduces computational overhead while mitigating the performance degradation caused by BFP16 storage. Our approach applies broadly to Shampoo-based methods that employ QR decomposition, including KL-Shampoo, SOAP, and KL-SOAP. In particular, it improves the performance of SOAP and KL-SOAP under BFP16 storage, enabling KL-SOAP to match or exceed KL-Shampoo. Overall, our approach makes Shampoo-based methods more memory- and time-efficient.

\end{abstract}

\section{Introduction}

\begin{figure}
    \centering
    \begin{minipage}[c]{0.5\linewidth}
      \centering
      \includegraphics{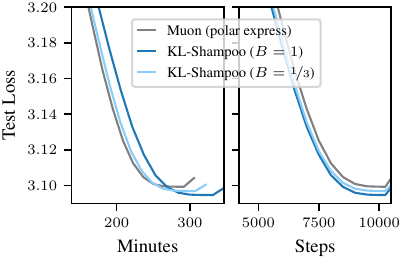}
    \end{minipage}\hfill
    \begin{minipage}[c]{0.5\linewidth}
      \centering
      \includegraphics{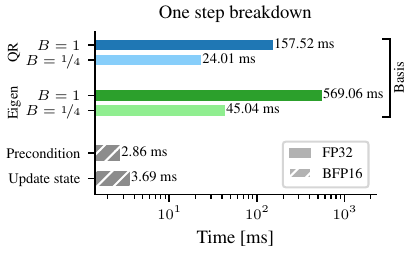}
    \end{minipage}

    \vspace{0.5em}

    \begin{minipage}[c]{0.48\linewidth}
      \raggedright
      \includegraphics{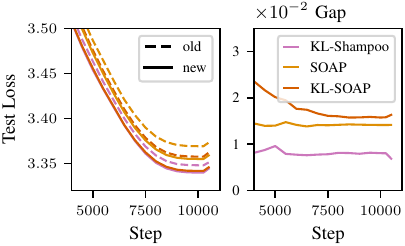}
    \end{minipage}\hfill
    \begin{minipage}[c]{0.5\linewidth}
\caption{\textbf{Top left:} Performance of full-basis and subspace KL-Shampoo (with BFP16 storage) on Llama3 (598M) trained on FineWeb 4.1B, evaluated using the hyperparameters tuned for full-basis KL-Shampoo over 120 runs.  Muon is also tuned independently over 120 runs. 
\textbf{Top right:} Computational cost for a weight matrix in $\mathcal{R}^{d_1 \times d_2}$ with $d_1 \!=\! d_2 \!= \! 6000$, and in a subspace with $d_\text{sub}\!=\!B d_1$, on an H100. \textbf{Bottom:} Test loss and per-method gap (old\,$-$\,new) on Llama3 (119M) trained on FineWeb 4.1B for KL-Shampoo, SOAP, and KL-SOAP, showing that the reparametrized variants outperform the original baselines when using BFP16 storage and full-basis updates. This gap highlights the performance degradation caused by BFP16 storage under the original parametrization.
      \vspace{-0.4cm}
      }
      \label{fig:time_cost}
    \end{minipage}
\end{figure}

Shampoo-based methods, such as Shampoo \citep{gupta18shampoo,shi2023distributed}, SOAP \citep{vyas2024soap}, and KL-Shampoo/SOAP \citep{lin2026understanding}, have recently attracted considerable attention because of their strong performance in training neural networks \citep{dahl2023benchmarking,kasimbegaccelerating,eschenhagen2026clarifying}. They employ non-diagonal preconditioners and, for a matrix-valued weight, use a Kronecker-factored preconditioner together with its inverse matrix root for preconditioning. To compute this root, pure Shampoo-based methods often perform eigendecompositions of each Kronecker factor, $\mS_i = \mQ_i \mathrm{Diag}(\vlambda_i)\mQ_i^\top$, and store ${\vlambda_i, \mQ_i, \mS_i}$ \citep{shi2023distributed,eschenhagen2025purifying}. SOAP-type methods also rely on eigendecomposition because they run Adam in the eigenbasis $\mQ_i$. However, eigendecomposition becomes expensive for large matrices, and existing implementations require single-precision (FP32)  arithmetic (e.g., LAPACK \citep{anderson1999lapack} on CPUs and cuSOLVER/MAGMA \citep{nvidia_cusolver,magma} on GPUs, as used in JAX \citep{bradbury2018jax} and PyTorch \citep{paszke2019pytorch}). As a result, preconditioning factors such as $\mQ_i$ and $\mS_i$ are typically stored in single precision \citep{anil2020scalable,shi2023distributed,eschenhagen2025purifying}. More recently, QR decomposition has been proposed as a cheaper approximation to eigendecomposition, reducing computational cost while retaining the same arithmetic \citep{vyas2024soap,eschenhagen2025purifying,lin2026understanding}. Still, QR decomposition remains the main computational bottleneck and is expensive relative to other subroutines (e.g., matrix multiplications; see \cref{fig:time_cost}, right). One possible way to reduce this cost is to consider subspace updates. However, it remains unclear how to do so using the current optimizer state. Furthermore, even when QR decomposition is performed in single precision, maintaining the optimizer state in half precision (BFP16) to reduce memory consumption can degrade performance.

In this work, we propose a reparametrization of the preconditioning factors in Shampoo-based methods to address these limitations.
Specifically, for each Kronecker factor $\mS_i$, we store $\{\vlambda_i, \mQ_i, \mP_i\}$ instead of $\{\vlambda_i, \mQ_i, \mS_i\}$, where $\mP_i := \mQ_i^\top \mS_i \mQ_i$, and we directly update $\mP_i$ without materializing $\mS_i$:
\begin{itemize}
    \item This enables efficient updates of a subset of the (orthogonal) basis vectors in $\mQ_i$ via QR in subspaces of $\mP_i$, thereby significantly reducing the cost of QR (see the top right of \cref{fig:time_cost}) while supporting memory-efficient half-precision storage.
    Unlike the existing parametrization, our reparametrization allows us to combine the newly updated basis vectors with the old ones to form a complete eigenbasis $\mQ_i$ and unifies full-basis and subspace updates.

    \item It is compatible with a variety of subspace selection strategies, including random selection and Jacobi-style selection \citep{van1985block,yamamoto2014convergence} originally designed for subspace eigendecomposition.
    This approach applies broadly to Shampoo-based methods that employ QR decomposition, including  KL-Shampoo, SOAP, and KL-SOAP.

    \item Empirically, our approach improves the performance of SOAP-type methods and closes the performance gap between KL-Shampoo and KL-SOAP reported by \citet{lin2026understanding} (see bottom left of \cref{fig:time_cost} and \cref{tab:support_bfp16}), even with full-basis QR updates, when all preconditioning factors are stored in BFP16.
    Our approach narrows the runtime gap between KL-Shampoo and Muon and opens new directions to improve Shampoo-based methods (see \cref{app:future}).
\end{itemize}

\vspace{-0.2cm}
\section{Background}
\label{sec:background}
\paragraph{Notation}
Because each weight matrix is treated independently in Shampoo-based methods, we focus on a single weight matrix, denoted by $\mTheta \in \sR^{d_1 \times d_2}$, rather than on all weight matrices in neural network optimization, in order to simplify the notation. To further simplify the discussion, we omit weight decay and momentum, which are also used in Shampoo-based methods. Here, $\mG$ denotes the gradient matrix with respect to $\mTheta$, and $\vg := \mathrm{vec}(\mG) \in \sR^{d_1 d_2 \times 1}$ denotes the flattened form of $\mG$. We use $\mathrm{diag}(\cdot)$ to extract the diagonal entries of an input matrix, while $\mathrm{Diag}(\cdot)$ denotes the diagonal matrix whose diagonal entries are given by the input vector.

\paragraph{Pure Shampoo-based methods}
These methods, such as Shampoo \citep{shi2023distributed,morwani2024new,eschenhagen2025purifying} and KL-Shampoo \citep{lin2026understanding}, employ a Kronecker-factorized preconditioning matrix $\mS_\text{shampoo}=\mS_1 \otimes \mS_2$ for each matrix-valued weight $\mTheta$ and perform the following preconditioning step using an inverse matrix square root:
\begin{align*}
   \vtheta &\leftarrow \vtheta - \gamma \mS_\text{shampoo}^{-\nicefrac{1}{2}}\, \vg \iff \mTheta \leftarrow \mTheta - \gamma \mS_1^{-\nicefrac{1}{2}} \mG \mS_2^{-\nicefrac{1}{2}},
\end{align*}
where  $\vtheta:=\mathrm{vec}(\mTheta)$,  $\otimes$ denotes a Kronecker product, $\gamma > 0$ is the step size, $\mS_i \in \sR^{d_i \times d_i}$ is a Kronecker factor matrix, and the equivalence follows from properties of the Kronecker product.

\paragraph{Eigendecomposition simplifies the computation of the matrix root}
Many existing works \citep{shi2023distributed,eschenhagen2025purifying} suggest performing an eigendecomposition of each Kronecker factor, $\mS_i=\mQ_i \mathrm{Diag}(\vlambda_i) \mQ_i^\top$ for $i \in \{1,2\}$, in order to compute the inverse square root. This yields
\begin{align*}
\mS_\text{shampoo}^{-\nicefrac{1}{2}}
&= (\mQ_1 \mathrm{Diag}(\vlambda_1^{\odot -\nicefrac{1}{2}}) \mQ_1^\top) \otimes (\mQ_2 \mathrm{Diag}(\vlambda_2^{\odot -\nicefrac{1}{2}}) \mQ_2^\top) 
= \mQ \mathrm{Diag}(\vlambda_1^{\odot -\nicefrac{1}{2}} \otimes \vlambda_2^{\odot -\nicefrac{1}{2}}) \mQ^\top,
\end{align*}
where $\odot$ denotes an elementwise operation, $\mQ_i$ is the orthogonal factor obtained from the decomposition, and $\mQ := \mQ_1 \otimes \mQ_2$ is the eigenbasis of $\mS_\text{shampoo}$.

\paragraph{SOAP-type methods}
Methods such as SOAP \citep{vyas2024soap} and KL-SOAP \citep{lin2026understanding} require eigendecomposition or its approximations because they employ an augmented matrix
$\mS_{\text{soap}} := \mQ \mathrm{Diag}(\vd) \mQ^\top$
for preconditioning, where $\vd \in \sR^{d_1 d_2 \times 1}$ is known as Adam's second-moment vector in the basis $\mQ$. Notably, $\vd$ cannot be expressed as a Kronecker product, such as $\vlambda_1 \otimes \vlambda_2$, as in pure Shampoo-based methods \citep{lin2026understanding}. Thus, a SOAP-type method requires eigendecomposition or its approximations in order to estimate $\vd$.

\paragraph{Using stale eigenbasis can reduce the runtime cost with the price of performance degradation}
Many existing works perform eigendecomposition infrequently because of its high computational cost (see \cref{fig:time_cost}). The eigenbasis $\mQ$ obtained from the most recent decomposition is stored and reused for preconditioning in order to avoid performing the decomposition at every iteration. To make Shampoo-based methods competitive, the decomposition interval $T$ must be large (e.g., performing the decomposition every $T=50$ gradient steps). However, a large decomposition interval can significantly degrade the performance of Shampoo-based methods \citep{shi2023distributed,vyas2024soap,eschenhagen2025purifying} because of the staleness of the eigenbasis. Thus, we cannot simply reduce the cost of this decomposition by increasing the interval without sacrificing performance.

\paragraph{QR decomposition is a cheaper approximation
to improve runtime and reduce performance degradation}
Motivated by the trade-off between performance and computational cost, QR decomposition has been proposed as a cheaper approximation to eigendecomposition \citep{vyas2024soap,eschenhagen2025purifying,lin2026understanding}: $\smash{\mQ_i^\text{(new)} = \mathrm{qr}(\mS_i \mQ_i^\text{(old)}) \text{ for } i \in \{1, 2\}}$, to reduce computational cost (see \cref{fig:time_cost}) while maintaining the performance of Shampoo-based methods, where $\smash{\mQ_i^\text{(old)}}$ is a stale eigenbasis. This is possible because QR decomposition allows a smaller decomposition interval (e.g., $T=10$) to improve the performance while keeping the cost low.
\paragraph{Tracking eigenvalues is essential when using stale bases}
Because QR decomposition can only approximate eigenbases, existing works propose tracking eigenvalues separately, for example, through estimation schemes for pure Shampoo-based methods \citep{lin2026understanding} and SOAP-type methods \citep{vyas2024soap}. \citet{eschenhagen2025purifying,lin2026understanding} further show that updating eigenvalues at each step is essential when using outdated bases $\mQ_i$ to reduce runtime cost. We will show our reparametrization is compatible with these schemes (see Step 3 of \cref{box:qr_for_kl}).

\subsection{Limitations induced by matrix decomposition}

As we discussed before, matrix decomposition is required by SOAP-type methods and recommended for pure Shampoo-based methods. However, there are limitations in using matrix decomposition that must be overcome to further unlock the potential of these methods for neural network optimization.

\paragraph{Half-precision may degrade performance, whereas single precision increases memory consumption}
Existing implementations of eigendecomposition and QR decomposition in LAPACK and cuSOLVER/MAGMA, as used in JAX and PyTorch, require single-precision arithmetic in order to remain numerically stable. This is because these algorithms are typically studied and analyzed in single-precision and, in some cases, double-precision settings \citep{golub2013matrix}. Decomposition algorithms that support half-precision computation are not yet widely available and remain under active development \citep{higham2022mixed}. As a result, preconditioning factor matrices are often stored in full precision \citep{anil2020scalable,shi2023distributed}. Unfortunately, this increases memory consumption. Although mixed-precision schemes \citep{vyas2024soap,lin2026understanding}, such as combining half-precision storage with full-precision decomposition, are possible, half-precision storage can degrade the performance of Shampoo-based methods. For example, \citet{lin2026understanding} report that KL-SOAP underperforms KL-Shampoo when BFP16 storage is used.

\paragraph{Matrix decomposition is the dominant computational cost}
Matrix decomposition algorithms such as eigen- and QR decomposition have cubic complexity and, unlike matrix multiplication, they require full-precision arithmetic and are difficult to parallelize on GPUs. \Cref{fig:time_cost} shows that matrix decompositions are main computational bottleneck in Shampoo-based methods. Thus, making QR decomposition faster can make these methods more competitive. %

\providecommand{\cost}[1]{\textcolor{gray}{\# \ensuremath{O(#1)}}}
\providecommand{\mms}[1]{\textcolor{gray}{\# \ensuremath{#1}\,\texttt{\ifnum#1=1 mm\else mms\fi}}}
\providecommand{\smms}[1]{\textcolor{gray}{\# \ensuremath{#1}\,\texttt{\ifnum#1=1 smm\else smms\fi}}}

\begin{figure*}[!t]
  {\setlength{\fboxsep}{0pt}\fbox{%
  \begin{minipage}[t]{\dimexpr\textwidth-2\fboxrule\relax}
    \begin{tabular}{@{}>{\centering\arraybackslash}p{0.49\linewidth}@{\hspace{0.01\linewidth}}|@{\hspace{0.01\linewidth}}>{\centering\arraybackslash}p{0.49\linewidth}@{}}
      \textbf{KL-Shampoo (Old)} &
      \textbf{KL-Shampoo \highlight{(Reparameterized)}} \\[0.3em]

      Optimizer state &
      Optimizer state \\
      $\{\vlambda_i, \mQ_i, \mS_i \mid i = 1, 2\}$ &
      $\{\vlambda_i, \mQ_i, \highlight{\mP_i}: = \mQ_i^\top \mS_i \mQ_i \mid i = 1, 2\}$ \\
      \midrule

        \multicolumn{2}{@{}>{\centering\arraybackslash}p{0.97\linewidth}@{}}{1:\ Compute gradient $\vg := \nabla \ell(\vtheta)$,\ $\mG := \Mat(\vg) \in \sR^{d_1 \times d_2}$} \\[0.4em]
        \midrule

        \multicolumn{2}{@{}>{\centering\arraybackslash}p{0.97\linewidth}@{}}{2:\ Estimate preconditioning factors (each iteration, full-basis)} \\
        $\begin{array}[t]{r@{\;}c@{\;}l@{\quad}l}
          & & & \\
          \mG'_1 & := & \mG \mQ_2 \Diag(\vlambda_2^{-\nicefrac{1}{2}}) / \sqrt{d_2} & \text{\mms{1}} \\
          \mG'_2 & := & \mG^\top \mQ_1 \Diag(\vlambda_1^{-\nicefrac{1}{2}}) / \sqrt{d_1} & \text{\mms{1}} \\
          \Delta_i & := & \mG'_i \mG_i^{\prime\top} & \text{\mms{2}} \\[0.4em]
          \mS_i & \leftarrow & (1 - \beta_2) \mS_i + \beta_2 \Delta_i &
        \end{array}$ &
        $\begin{array}[t]{r@{\;}c@{\;}l@{\quad}l}
          \Tilde{\mG}' & := & \mQ_1^\top \mG \mQ_2 & \text{\mms{2}} \\
          \highlightii{\Tilde{\mG}'_1} & := & \Tilde{\mG}' \Diag(\vlambda_2^{-\nicefrac{1}{2}}) / \sqrt{d_2} & \\
          \highlightii{\Tilde{\mG}'_2} & := & \Tilde{\mG}^{\prime\top} \Diag(\vlambda_1^{-\nicefrac{1}{2}}) / \sqrt{d_1} & \\
          \Tilde{\Delta}_i & := & \highlightii{\Tilde{\mG}'_i} \highlightii{\Tilde{\mG}_i^{\prime\top}} \equiv \mQ_i^\top \Delta_i \mQ_i & \text{\mms{2}} \\[0.4em]
          \highlight{\mP_i} & \leftarrow & (1 - \beta_2) \highlight{\mP_i} + \beta_2 \Tilde{\Delta}_i &
        \end{array}$ \\
        \midrule

        \multicolumn{2}{@{}>{\centering\arraybackslash}p{0.97\linewidth}@{}}{3:\ Track eigenvalues via EMA (each iteration, full-basis)} \\
        $\begin{array}[t]{r@{\;}c@{\;}l@{\quad}l}
          \highlightii{\Tilde{\mG}'_i} & := & \mQ_i^\top \mG'_i & \text{\mms{2}} \\
          \diag(\Tilde{\Delta}_i) & = & (\highlightii{\Tilde{\mG}'_i} \odot \highlightii{\Tilde{\mG}'_i}) \mathbf{1} &
        \end{array}$ &
        Use $\Tilde{\Delta}_i$ for free \\
        \multicolumn{2}{@{}c@{}}{$\vlambda_i \leftarrow (1 - \beta_2)\vlambda_i + \beta_2 \diag(\Tilde{\Delta}_i)$} \\[0.4em]
        \midrule

        \multicolumn{2}{@{}>{\centering\arraybackslash}p{0.97\linewidth}@{}}{4:\ Estimate eigenbasis via QR (every $T \geq 1$ iterations, full-basis or subspace)} \\
        $\begin{array}[t]{r@{\,}c@{\,}l@{\quad}l}
          \mZ_i & \leftarrow & \mS_i \mQ_i & \text{\mms{2}} \\
          \mQ_i & \leftarrow & \qr(\mZ_i) & \text{\cost{d_i^3}}
        \end{array}$ &
        $\begin{array}[t]{r@{\,}c@{\,}l@{\quad}l}
          \mO_i & \leftarrow & \qr(\highlight{\mP_i}) & \text{\cost{d_i^3}} \\
          \mQ_i & \leftarrow & \mQ_i \mO_i & \text{\mms{2}} \\
          \highlight{\mP_i} & \leftarrow & \mO_i^\top \highlight{\mP_i} \mO_i & \text{\mms{4}}
        \end{array}$ \\
        \midrule

      \multicolumn{2}{@{}>{\centering\arraybackslash}p{0.97\linewidth}@{}}{5:\ Precondition using $\mQ := \mQ_1 \otimes \mQ_2$ and learning rate $\gamma$} \\
      \multicolumn{2}{@{}c@{}}{$\vtheta \leftarrow \vtheta - \gamma ( \mQ \Diag(\vlambda_1 \otimes \vlambda_2)^{-\nicefrac{1}{2}} \mQ^\top ) \vg$ \quad\cost{d_1^2 d_2 + d_1 d_2^2}} \\
    \end{tabular}
  \end{minipage}
  }}
  \caption{\textbf{Side-by-side comparison of KL-Shampoo (left) and its reparameterized variant (right).} The reparameterized variant stores $\smash{\highlight{\mP_i} := \mQ_i^\top \mS_i \mQ_i}$ instead of $\mS_i$; given equivalent initial state, the two variants produce equivalent iterates (see \cref{sec:full_basis}), justifying the term \emph{reparameterization}.
  \textbf{Color code.} Symbols in \highlight{red} mark where the two variants differ. Symbols in \highlightii{blue} mark the rotated factors $\smash{\highlightii{\Tilde{\mG}'_i} = \mQ_i^\top \mG'_i}$: the old variant forms them in step 3 to extract $\smash{\diag(\Tilde{\Delta}_i)}$, while the reparameterized variant obtains them from the shared intermediate $\smash{\Tilde{\mG}' = \mQ_1^\top \mG \mQ_2}$ in step 2 and reuses them to update $\highlight{\mP_i}$ --- thereby reusing work that is already done.
  \textbf{Cost.} Gray annotations count matrix--matrix products (\texttt{mm}), summed over $i = 1, 2$. The covariance/eigenvalue update (steps 2--3, every iteration) is $2\,\texttt{mm}$ cheaper for the reparameterized variant, whereas the eigenbasis update (step 4, every $T \geq 1$ iterations) costs $4\,\texttt{mm}$ more; the latter amortizes for any $T \geq 2$, so the reparameterization is strictly cheaper in the regime $T \geq 2$ used in practice.}
  \label{box:qr_for_kl}
\end{figure*}

\vspace{-0.2cm}
\section{Reparametrization for Reducing Time and Memory Costs}
We propose to reduce time and memory costs by reparametrizing the preconditioning factors $\mS_i$ as $\mP_i=\mQ_i^\top \mS_i \mQ_i$. This reparametrization enables Shampoo-based methods to support both full-basis and subspace orthogonal updates, as well as half-precision storage, while empirically mitigating performance degradation.
As we will show, these updates require performing QR decomposition on a subspace of $\mP_i$ (see \cref{box:subspace_wip}).
This motivates directly storing and using $\mP_i$ as the reparametrization of the preconditioning factor $\mS_i$.

We begin by modifying an existing update scheme to use this reparametrization. To this end, we establish the mathematical equivalence between the existing parametrization (e.g., $\mS_i$) and our reparametrization (e.g., $\mP_i$) in the full-basis setting. We then show that, unlike the existing parametrization, our reparametrization not only supports BFP16 storage without compromising performance in the full-basis setting, but also enables efficient subspace orthogonal updates (e.g., updates of a subset of basis vectors in $\mQ_i$ via QR on a subspace of $\mP_i$). Notably, we keep the updates of the eigenvalues (e.g., $\vlambda_i$ for pure Shampoo-based methods or $\vd$ for SOAP-type methods) and preconditioning factors (e.g., $\mP_i$) in the full basis (see Step 2 \& 3 in  \cref{box:qr_for_kl}). This is essential for mitigating performance degradation of these methods under subspace updates while reducing the cost of QR decomposition, the main computational bottleneck (see the top-right panel of \cref{fig:time_cost}).

\subsection{Full-basis Update: Memory Reduction via Half-precision Storage}
\label{sec:full_basis}

For simplicity, we assume in this section that a QR implementation returns a unique decomposition. 
See \cref{app:qr_unique} for making any QR implementation unique.

Recall that existing QR-based Shampoo methods \citep{vyas2024soap, lin2026understanding} store $\mS_i$ and compute $\smash{\mQ_i^\text{(new)}}$ via a full-basis QR decomposition $\smash{\mQ_i^\text{(new)} \mR_i^\text{(new)} = \mS_i \mQ_i^\text{(old)}}$ to approximate an orthogonal eigenbasis of $\mS_i$, for $i \in \{1,2\}$, For example, consider the KL-Shampoo update in \cref{box:qr_for_kl}.

\paragraph{Mathematical equivalence for deriving our update scheme}
Modifying the existing QR-based update scheme for our reparametrization builds on the observation that performing QR decomposition on the reparametrized matrix
$\mP_i^\text{(old)} := \mQ_i^{\text{(old)}\top} \mS_i \mQ_i^\text{(old)}$
yields
\begin{align}
\mO_i \bar{\mR}_i
    \!=\! 
    \mP_i^\text{(old)} \!=\!
    \mQ_i^{\text{(old)}\top} \big[ \mS_i \mQ_i^\text{(old)} \big] 
    \!= \!\mQ_i^{\text{(old)}\top} \big[ \mQ_i^\text{(new)} \mR_i^\text{(new)} \big] \!=\! \big[
\mQ_i^{\text{(old)}\top} \mQ_i^\text{(new)} \big]\mR_i^\text{(new)}
  \label{eq:qr_new}
\end{align}
We then use the equivalence established in \cref{eq:qr_new} to obtain a new update rule for computing $\mQ_i$ under our reparametrization. Notice that both the left-hand side and the right-hand side of \cref{eq:qr_new} are QR decompositions of $\smash{\mP_i^\text{(old)}}$.
Because the QR decomposition for non-singular $\smash{\mP_i^\text{(old)}}$ is unique under our assumption, we obtain the relationship
\begin{align*}
     \mO_i = \mQ_i^{\text{(old)}\top} \mQ_i^\text{(new)}, \qquad \bar{\mR}_i = \mR_i^\text{(new)}.
\end{align*}
This implies that $\mQ_i^\text{(new)} = \mQ_i^\text{(old)} \mO_i$ because $\mQ_i^\text{(old)} \mQ_i^{\text{(old)}\top} = \mI$.
This relationship leads to the following update scheme for $\mQ_i$ under our reparametrization: perform QR decomposition on $\smash{\mP_i^\text{(old)}}$ to obtain $\mO_i$, and then update $\mQ_i$ via
\begin{align}
    \mQ_i^\text{(new)} = \mQ_i^\text{(old)} \mO_i.
    \label{eq:full_new_qr}
\end{align} 
Importantly, this scheme enables efficient subspace update schemes, as will be discussed in \cref{sec:subspace_qr}.

Since $\mP_i$ depends on $\mQ_i$, we need to update $\mP_i$ via
\begin{align}
    \mP_i^\text{(new)}
    :=
    \mQ_i^{\text{(new)}\top} \mS_i \mQ_i^\text{(new)}
    =
    \mO_i^\top \mQ_i^{\text{(old)}\top} \mS_i \mQ_i^\text{(old)} \mO_i
    =
    \mO_i^\top \mP_i^\text{(old)} \mO_i,
    \label{eq:full_new_P}
\end{align}
which avoids explicitly forming $\mS_i$ when the basis is changed from $\mQ_i^\text{(old)}$ to $\mQ_i^\text{(new)}$.

If $\mQ_i$ is held fixed while $\mS_i$ is updated via an exponential moving average (EMA) with $\Delta_i$
shown in  Step 2 in \cref{box:qr_for_kl}, then $\mP_i$ is equivalently updated using $\smash{\Tilde{\Delta}_i := \mQ_i^\top \Delta_i \mQ_i}$:
\begin{align}
\mQ_i^\top \mS_i^\text{(new)} \mQ_i
= (1-\beta_2)\mQ_i^\top \mS_i^\text{(old)} \mQ_i + \beta_2 \mQ_i^\top\Delta_i \mQ_i
\iff \mP_i^\text{(new)} =
(1-\beta_2)\mP_i^\text{(old)} + \beta_2 \Tilde{\Delta}_i.
\end{align}

\paragraph{Eigenvalue estimation and preconditioning remain unchanged under our reparametrization}
The existing schemes for eigenvalue estimation and preconditioning require only knowledge of $\mQ_i$, rather than $\mS_i$ (see Step 3 in \cref{box:qr_for_kl}). Thus, our reparametrization is compatible with these schemes.

\paragraph{Generalization to other Shampoo-based methods}
Although we describe the changes for KL-Shampoo in \cref{box:qr_for_kl}, our approach applies directly to other Shampoo-based methods that use QR decomposition. Pure Shampoo-based methods often differ in how they compute $\Delta_i$ in Step 2 of \cref{box:qr_for_kl}. Thus, our approach applies to them. SOAP-type methods do not require Step 3, but they do require a further modification of Step 5 in \cref{box:qr_for_kl} to run Adam in the basis $\mQ_i$, which requires only knowledge of $\mQ_i$ rather than $\mS_i$. Therefore, our reparametrization also applies to SOAP-type methods.

\paragraph{Our reparametrization supports half-precision storage and encourages sparsity}
While the existing parametrization and our reparametrization are mathematically equivalent, our empirical results (see \cref{tab:support_bfp16}) show that our reparametrization, unlike the old parametrization, preserves the performance of Shampoo-based methods under half-precision storage, thereby enabling half-precision storage without compromising performance. 
Moreover, the preconditioning factor $\mP_i$ is often close to a diagonal matrix and contains many near-zero off-diagonal entries when $\mQ_i$ is updated frequently. This is because $\mQ_i$ approximates the eigenbasis of $\mS_i$, and consequently, $\mP_i := \mQ_i^\top \mS_i \mQ_i$ becomes nearly diagonal. 
This sparsity can be leveraged to further reduce memory usage.

\providecommand{\cost}[1]{\textcolor{gray}{\# \ensuremath{O(#1)}}}
\providecommand{\mms}[1]{\textcolor{gray}{\# \ensuremath{#1}\,\texttt{\ifnum#1=1 mm\else mms\fi}}}
\providecommand{\smms}[1]{\textcolor{gray}{\# \ensuremath{#1}\,\texttt{\ifnum#1=1 smm\else smms\fi}}}

\begin{figure*}[!t]
  \begin{minipage}[c]{0.76\linewidth}
  {\setlength{\fboxsep}{0pt}\fbox{%
  \begin{minipage}[t]{\dimexpr\linewidth-2\fboxrule\relax}
    \begin{tabular}{@{}>{\centering\arraybackslash}p{0.49\linewidth}@{\hspace{0.01\linewidth}}|@{\hspace{0.01\linewidth}}>{\centering\arraybackslash}p{0.49\linewidth}@{}}
      \textbf{KL-Shampoo (Old)} &
      \textbf{KL-Shampoo \highlight{(Reparameterized)}} \\[0.3em]

      \multicolumn{2}{@{}>{\centering\arraybackslash}p{0.97\linewidth}@{}}{4:\ Estimate eigenbasis via subspace QR (every $T \geq 1$ iterations)} \\
      \midrule
      \multicolumn{2}{@{}>{\centering\arraybackslash}p{0.97\linewidth}@{}}{4a:\ Select index set $\sI_i$ of $d_{\text{sub},i}$ columns of $\highlight{\mP_i}$} \\
      \multicolumn{2}{@{}>{\centering\arraybackslash}p{0.97\linewidth}@{}}{(i.e.\ $|\sI_i| =  d_{\text{sub},i} = B d_i $, $0<B<1$)} \\
     Compute $\highlight{\mP_i} = \mQ_i^\top \mS_i \mQ_i$ \quad \mms{4} & Use $\highlight{\mP_i}$ for free
      \\
      \multicolumn{2}{@{}c@{}}{$\sI_i := \textrm{select}(\highlight{\mP_i})$} \\
      \midrule
      \multicolumn{2}{@{}>{\centering\arraybackslash}p{0.97\linewidth}@{}}{4b:\ Perform subspace QR on $\highlight{\mP_i}$} \\
      \multicolumn{2}{@{}c@{}}{$\mO_i \leftarrow \qr(\highlight{\mP_i}[\sI_i, \sI_i])$ \quad \cost{d_{\text{sub},i}^3}} \\
      \midrule
      \multicolumn{2}{@{}>{\centering\arraybackslash}p{0.97\linewidth}@{}}{4c:\ Update only the $d_{\text{sub},k}$ orthogonal bases} \\
      \multicolumn{2}{@{}c@{}}{$\mQ_i[:, \sI_i] \leftarrow \mQ_i[:, \sI_i] \mO_i$ \quad \smms{2}} \\
      &
      \multicolumn{1}{@{\hspace{0.01\linewidth}}>{\raggedright\arraybackslash}p{0.49\linewidth}@{}}{\quad\quad Rotate $\highlight{\mP_i}$ in subspace:} \\
      &
      $\begin{array}[t]{r@{\;}c@{\;}l@{\quad}l}
        \highlight{\mP_i}[:, \sI_i] & \leftarrow & \highlight{\mP_i}[:, \sI_i] \mO_i & \text{\smms{2}} \\
        \highlight{\mP_i}[\sI_i, :] & \leftarrow & \mO_i^\top \highlight{\mP_i}[\sI_i, :] & \text{\smms{2}}
      \end{array}$ \\
    \end{tabular}
  \end{minipage}
  }}
  \end{minipage}\hfill
  \begin{minipage}[c]{0.195\linewidth}
    \caption{\textbf{Step 4 with subspace QR: side-by-side comparison.} Drop-in replacement for Step 4 which uses subspace QR. The reparameterization avoids the cost of forming $\highlight{\mP_i}$ at every QR step, since $\highlight{\mP_i}$ stays up-to-date.
    \textbf{Cost.} An \texttt{smm} (subspace \texttt{mm}) costs $B^2$ of an \texttt{mm}.}
    \label{box:subspace_wip}
  \end{minipage}
\end{figure*}

\subsection{Subspace Update: Computational Cost Reduction via QR Decomposition in Subspaces}
\label{sec:subspace_qr}
Our update scheme under the reparametrization directly supports updating a subset of the orthogonal basis vectors in $\mQ_i$ via QR in subspaces of $\mP_i$. Interestingly, the scheme in \cref{eq:full_new_qr,eq:full_new_P} resembles block Jacobi-type methods \citep{van1985block,yamamoto2014convergence} for subspace approximation via eigendecomposition. Thus, our reparametrization supports efficient subspace updates via either QR or eigendecomposition. In this work, however, we focus on a QR-based local orthogonal factor within a selected subspace because subspace QR is computationally cheaper (see the top right of \cref{fig:time_cost}) while empirically achieving performance similar to that of subspace eigendecomposition for Shampoo-based methods (see \cref{tab:exp3}).

We now describe how our reparametrization directly enables a single subspace update step. For illustration, suppose that $\mP_i$ is partitioned as
\begin{align*}
    \mP_i =
    \begin{bmatrix}
      \highlight{ \mP_i^{(XX)} }& \mP_i^{(XY)} \\
       \mP_i^{(YX)} & \mP_i^{(YY)}
    \end{bmatrix}   \quad \text{with } \mP_i[:, \sI_i]=    \begin{bmatrix}
      \highlight{ \mP_i^{(XX)} } \\
       \mP_i^{(YX)}
    \end{bmatrix},
\end{align*}
and that the block $\mP_i^{(XX)}=\mP_i[\sI_i, \sI_i]$, highlighted in red, is selected for the update. 

Conceptually, we perform QR decomposition only on $\smash{\mP_i^{(XX)}}$ to obtain a local orthogonal factor $\smash{\mO_i^{(XX)}}$, and then define $\smash{\mO_i = \mO_i^{(XX)} \oplus \mI^{(YY)}}$, where 
$\oplus$ represents the direct sum of matrices to form a block diagonal matrix and 
$\smash{\mI^{(YY)}}$ is an identity matrix. We then use this block-diagonal orthogonal matrix in the full-basis update rules shown in \cref{eq:full_new_qr,eq:full_new_P}. Substituting this form of $\mO_i$ into \cref{eq:full_new_qr} yields the following subspace update $\mQ_i^\text{(new)} = \mQ_i^\text{(old)} \mO_i =$
\begin{tcolorbox}[enhanced,colback=white,%
    colframe=red!75!black, attach boxed title to top right={yshift=-\tcboxedtitleheight/2, xshift=-1.25cm}, title=Partially Updating Orthogonal Basis Vectors via Subspace QR, coltitle=red!75!black, boxed title style={size=small,colback=white,opacityback=1, opacityframe=0}, size=title, enlarge top initially by=-\tcboxedtitleheight/2]
\vskip-0.5em
\begin{equation}
\resizebox{0.95\linewidth}{!}{$
   \begin{bNiceMatrix}[margin]
      (\mQ_i^\text{(old)})^{(XX)} & (\mQ_i^\text{(old)})^{(XY)} \\[4pt]
      (\mQ_i^\text{(old)})^{(YX)} & (\mQ_i^\text{(old)})^{(YY)}
      \CodeAfter
      \begin{tikzpicture}
        \draw[decorate, decoration={brace, amplitude=4pt, mirror}]
          ([yshift=-4pt]2-2.south west) -- ([yshift=-4pt]2-2.south east)
          node[midway, below=2pt] {\small\text{old basis}};
      \end{tikzpicture}
   \end{bNiceMatrix}
   \begin{bmatrix}
      \mO_i^{(XX)} & \mathbf{0} \\
      \mathbf{0} & \mI^{(YY)}
   \end{bmatrix}
   =
   \begin{bNiceMatrix}[margin]
      (\mQ_i^\text{(old)})^{(XX)} \mO_i^{(XX)} & (\mQ_i^\text{(old)})^{(XY)} \\[4pt]
      (\mQ_i^\text{(old)})^{(YX)} \mO_i^{(XX)} & (\mQ_i^\text{(old)})^{(YY)}
      \CodeAfter
      \begin{tikzpicture}
        \draw[decorate, decoration={brace, amplitude=4pt, mirror}]
          ([yshift=-4pt]2-1.south west) -- ([yshift=-4pt]2-1.south east)
          node[midway, below=2pt] {\small\text{new subspace basis}};
        \draw[decorate, decoration={brace, amplitude=4pt, mirror}]
          ([yshift=-4pt]2-2.south west) -- ([yshift=-4pt]2-2.south east)
          node[midway, below=2pt] {\small\text{unchanged basis}};
      \end{tikzpicture}
   \end{bNiceMatrix}.
$}
\label{eq:lowrank_Q}
\end{equation}
\vspace*{0.7em}
\end{tcolorbox}
Similarly,  we update $\mP_i$ in the subspace by substituting this form of $\mO_i$ into \cref{eq:full_new_P}
\begin{align}
    \mP_i^\text{(new)}  = \mO_i^\top \mP_i^\text{(old)} \mO_i
    =     \begin{bmatrix}
         \big(\mO_i^{(XX)}\big)^\top \mP_i^{(XX)}  \mO_i^{(XX)} & \big(\mO_i^{(XX)}\big)^\top  \mP_i^{(XY)} \\
       \mP_i^{(YX)} \mO_i^{(XX)}  & \mP_i^{(YY)}
    \end{bmatrix}
   \label{eq:lowrank_P}
\end{align}

Mathematically, $\mQ_i^\text{(new)}$ remains orthogonal because it is the product of two orthogonal matrices, and $\mO_i$ is orthogonal by construction.
The update in \cref{eq:lowrank_Q,eq:lowrank_P} recovers the block-Jacobi-type update \citep{yamamoto2014convergence} when using eigen-decomposition on $\smash{\mP_i^{(XX)}}$ rather than QR decomposition. Thus, our reparametrization supports subspace updates via either QR or eigen-decomposition.

\paragraph{Inner loop and early termination}
Block Jacobi methods approximate a fixed matrix $\mP_i$ by introducing an inner loop that repeatedly selects a block of $\mP_i$, computes a local orthogonal update $\mO_i$, and updates $\mQ_i$ through \cref{eq:full_new_qr}. Motivated by this idea, we can introduce the same $ K$-step loop into our scheme. Unlike block Jacobi-type methods, however, $\mP_i$ changes at each gradient step in Shampoo-based methods. This new setting allows us to terminate the loop early, trading approximation accuracy for runtime, since we do not need to approximate the ever-changing $\mP_i$ precisely. Importantly, this reduces the cost of QR by replacing a full-basis decomposition with a small number of subspace decompositions.
Empirically, using a single QR step in a subspace ($K=1$) is often sufficient for Shampoo-based methods when the decomposition interval is small (see \cref{tab:blocking2}).

\paragraph{Our reparametrization facilitates greedy block selections}

Both our approach and Jacobi-type methods use $\mQ_i$ to approximate an eigenbasis of $\mS_i$.
Like block Jacobi-type methods, our approach requires selecting a block of $\mP_i$ from which to compute a local orthogonal factor.
There is no closed-form solution for selecting an optimal block, and many greedy block-selection strategies have therefore been considered in the literature.
In the ideal case, $\mP_i = \mQ_i^\top \mS_i \mQ_i$ is diagonal.
Motivated by this observation, Jacobi-type greedy methods commonly use the Frobenius norm of the off-diagonal part of $\mP_i$, as a guide for selection.
Because our reparametrization stores $\mP_i$ explicitly, evaluating such objectives is inexpensive, whereas doing so under the existing parametrization (i.e., storing $\mS_i$) is more costly. Thus, our reparametrization makes it easy both to use existing greedy block-selection strategies and to design new greedy ones.

\paragraph{GPU-friendly greedy selection}
Existing block Jacobi-type selection methods \citep{bevcka2002dynamic,yamamoto2014convergence} require an additional loop to select a block. We instead consider a loop-free, GPU-friendly alternative: \textbf{a greedy two-phase method} inspired by the Jacobi method.
In phase 1, we select the classical greedy Jacobi pair \citep{forsythe1960cyclic},
\begin{align*}
   &\text{phase 1}:  \quad (k^*,j^*) = \arg \max_{k \neq j} P_{k,j}^2,
\end{align*}
and in phase 2, we expand this pair into a block of size $b$ by choosing the $b-2$ indices with the largest total coupling to $(k^*,j^*)$,
\begin{align*}
   &\text{phase 2}: \quad S^* = \arg\max_{\substack{S \subseteq [d]\setminus\{k^*,j^*\} \\ |S| = b-2}}
   \sum_{x \in S} \Big( P_{x,k^*}^2 + P_{x,j^*}^2 \Big).
\end{align*}
Here we drop the subscript of $\mP_i$, $b$ is the block size, $\sI_i := S^* \cup \{k^*, j^*\}$ is the selected index set, and $P_{k,j}$ denotes the $(k,j)$-th entry of $\mP_i$.
This can be implemented efficiently with \texttt{torch.topk}.

\vspace{-0.2cm}
\section{Experiments}
\label{sec:exps}
\begin{table}[t]
\centering
\caption{Performance of each method in terms of test loss. Each cell shows the result with FP32 storage followed by BFP16 storage; the number above the arrow is the (BFP16$-$FP32) delta. Small deltas indicate the parameterization is robust to half-precision storage. Best result per column (FP32 and BFP16 separately) is shown in bold. Our reparameterization is consistently best when switching from FP32 to BFP16 storage.}
\label{tab:support_bfp16}
\begin{tabular}{cccc}
\toprule
\textbf{Method} &
\makecell{$\!\!$\textbf{Parame-}$\!\!$ \\ $\!\!$\textbf{terization}$\!\!$}
  & \makecell{\textbf{nanoGPT (123M)} \\ FP32 $\to$ BFP16}
  & \makecell{\textbf{llama3 (119M)} \\ FP32 $\to$ BFP16}
  \\
\midrule
\multirow{2}{*}{KL-Shampoo}
& old
  & $3.239 \xrightarrow{+0.010} 3.249$
  & $\mathbf{3.345} \xrightarrow{+0.000} 3.345$
  \\
& new
  & $3.239 \xrightarrow{+0.005} 3.244$
  & $\mathbf{3.345} \xrightarrow{-0.001} \mathbf{3.344}$
  \\
\midrule
\multirow{2}{*}{KL-SOAP}
& old
  & $3.239 \xrightarrow{+0.013} 3.252$
  & $3.351 \xrightarrow{+0.011} 3.362$
  \\
& new
  & $\mathbf{3.235} \xrightarrow{+0.006} \mathbf{3.241}$
  & $3.346 \xrightarrow{+0.000} 3.346$
  \\
\midrule
\multirow{2}{*}{SOAP}
& old
  & $3.248 \xrightarrow{+0.013} 3.261$
  & $3.358 \xrightarrow{+0.014} 3.372$
  \\
& new
  & $3.247 \xrightarrow{+0.005} 3.252$
  & $3.353 \xrightarrow{+0.006} 3.359$
  \\
\bottomrule
\end{tabular}
\end{table}

We conduct five experiments to demonstrate the benefits of our reparametrization. In Experiment 1, we consider the Shampoo-based methods KL-Shampoo, SOAP, and KL-SOAP. The remaining experiments focus on KL-Shampoo as a representative Shampoo-based method because of limited computational resources. Our subspace approach directly applies to SOAP and KL-SOAP, too.

\paragraph{Experimental setup}
We consider the following baseline training methods: SOAP \citep{vyas2024soap}, KL-Shampoo/SOAP \citep{lin2026understanding}, and Muon \citep{liu2025muon}. We train language models---nanoGPT \citep{nanoGPT2024} (123M) and Llama 3 \citep{llama3} (119M, 313M, and 598M)---on the FineWeb dataset. We use the official implementations of SOAP \citep{vyas2024soap} and KL-Shampoo/SOAP \citep{lin2026understanding}. For Muon, we use the polar express implementation  \citep{amsel2026polar}.
For each baseline, we tune all available hyperparameters, including the learning rate, weight decay, damping, $\beta_1$, and $\beta_2$, using random search over 120 runs. For Shampoo-based methods, we set the decomposition interval to $T=10$, as suggested by \citet{vyas2024soap} and \citet{lin2026understanding}. In our experiments, the reparametrized methods (e.g., SOAP, KL-Shampoo, and KL-SOAP), including both full-basis and subspace variants, simply reuse the optimal hyperparameters found for the original parametrization via random search. By default, we use BFloat16 storage for each method. See \cref{app:exp_details} for additional experimental details.

\begin{table*}[t]
\begin{minipage}[t]{0.48\linewidth}
\centering
\caption{This demonstrates how subspace selection methods supported by our reparametrization can affect the performance of KL-Shampoo (ours, BFP16, QR) on test loss. As shown below, the greedy method is preferred over the random method due to its robustness on different models.}
\label{tab:exp2}
\begin{tabular}{cccccc}
\toprule

$T$ &
$B$ & $K$ & \textbf{Select} &
   \makecell{$\!\!$\textbf{nanoGPT}$\!\!$ \\ \textbf{(123M)}}
  & \makecell{\textbf{llama3} \\ \textbf{(119M)}}
  \\
\midrule
 \multirow{6}{*}{10} &  \multirow{6}{*}{$\nicefrac{1}{2}$} & \multirow{2}{*}{1}  & random
  & $3.249$
  & $3.345$
  \\
   &   &   & greedy
  & $3.247$
  &  $3.344$
  \\
  &  & \multirow{2}{*}{3} & random
  & $3.249$
  & $3.344$
  \\
    &  &  & greedy
  & $3.245$
  & $3.344$
  \\
 &   & \multirow{2}{*}{5}  & random
  & $3.249$
  &  $3.345$
  \\
&   &   &  greedy
  & $3.244$
  & $3.345$
  \\
\bottomrule
\end{tabular}
\end{minipage}\hfill
\begin{minipage}[t]{0.48\linewidth}
\centering
\caption{
This demonstrates that subspace orthogonal bases of
 KL-Shampoo (ours, BFP16, greedy) can be updated via an eigen- or QR-decomposition, as supported by our representation. As shown below, both decomposition methods perform similarly on test loss. This motivates the use of QR, as it is lower-cost.
}
\label{tab:exp3}
\begin{tabular}{cccccc}
\toprule

$T$ &
$B$ & $K$ & \textbf{Basis} &
   \makecell{$\!\!$\textbf{nanoGPT}$\!\!$ \\ \textbf{(123M)}}
  & \makecell{\textbf{llama3} \\ \textbf{(119M)}}
  \\
\midrule
 \multirow{6}{*}{10} &  \multirow{6}{*}{$\nicefrac{1}{2}$} & \multirow{2}{*}{1}  & Eig
  & $3.247$ %
  & $3.345$ %
  \\
   &   &   & QR
  & $3.247$ %
  &  $3.344$ %
  \\
  &  & \multirow{2}{*}{3} & Eig
  & $3.245$ %
  &  $3.344$ %
  \\
    &  &  & QR
  & $3.245$ %
  &  $3.344$ %
  \\
 &   & \multirow{2}{*}{5}  & Eig
  & $3.245$ %
  & $3.344$ %
  \\
&   &   & QR
  &  $3.244$ %
  &  $3.345$ %
  \\
\bottomrule
\end{tabular}
\end{minipage}

\end{table*}

\paragraph{Experiment 1: Our reparametrization preserves performance under BFloat16 storage}
In this set of experiments (see \cref{tab:support_bfp16}), we show that our reparametrization preserves the performance of Shampoo-based methods when switching the storage format from FP32 to BFP16, even in the full-basis setting. The existing parametrization does not maintain performance under BFP16 storage. According to \cref{tab:support_bfp16}, our reparametrization also closes the performance gap between KL-Shampoo and KL-SOAP reported by \citet{lin2026understanding} when BFP16 storage is used (see bottom-left of \cref{fig:time_cost}). 

\vspace{-0.1cm}
\paragraph{Experiment 2: Our reparametrization supports subspace selection methods}
This set of experiments (see \cref{tab:exp2}) illustrates the use of subspace selection methods with a $K$-step inner loop. A straightforward GPU-friendly block-selection strategy is random selection via uniform sampling. Thus, we consider both this random strategy and the greedy method described in \cref{sec:subspace_qr}. From \cref{tab:exp2}, we can see that the greedy selection is generally more effective than random selection. On nanoGPT, the greedy method performs much better because of nanoGPT's aggressive learning schedule. Based on these results, we focus on the greedy selection method in the remaining experiments.

\vspace{-0.1cm}
\paragraph{Experiment 3: Our reparametrization supports subspace updates via eigendecomposition or QR decomposition}
As discussed in \cref{sec:subspace_qr}, our QR-based update scheme coincides with block Jacobi methods. As a result, our reparametrization also supports subspace updates via eigendecomposition. In this set of experiments, we provide empirical evidence for this claim. These experiments also support the use of QR decomposition in a subspace. From \cref{tab:exp3}, we can see that eigendecomposition and QR decomposition perform similarly. However, QR decomposition is much faster than eigendecomposition in practice (see top right of  \cref{fig:time_cost}). This echoes the findings of \citet{vyas2024soap} and further motivates using QR  to replace eigendecomposition, even in subspaces.

\newcommand{\runtimeColorbarWorkshop}{%
\scalebox{0.8}{
  \begin{tikzpicture}[xscale=1.1]
    \foreach \i in {0,1,...,79} {%
      \pgfmathsetmacro{\rcR}{0.8 + \i/200}%
      \pgfmathsetmacro{\rcD}{abs(\rcR-1)}%
      \pgfmathtruncatemacro{\rcI}{\rcD*500}%
      \ifdim\rcR pt<1pt
        \fill[green!\rcI!yellow] (\i*0.05, 0) rectangle ({(\i+1)*0.05}, 0.3);
      \else
        \fill[red!\rcI!yellow] (\i*0.05, 0) rectangle ({(\i+1)*0.05}, 0.3);
      \fi
    }%
    \draw (0,0) rectangle (4,0.3);
    \foreach \x in {1, 2, 3}
      \draw[line width=0.6pt] (\x, 0) -- (\x, 0.3);
    \foreach \v/\x in {0.8/0, 0.9/1, 1.0/2, 1.1/3, 1.2/4}
      \node[below, font=\small] at (\x, 0) {\v};
    \node[left, xshift=-0.3em] at (0, 0.15) {runtime ratio};
  \end{tikzpicture}}%
}

\begin{table*}[t]
\def\baselineNanoGPT{209}%
\def\baselineNanoGPTStep{691.56}%
\def\baselineLlamaSmallStep{0.516}%
\def\baselineLlamaMid{1.1215}%
\begin{minipage}[t]{0.48\linewidth}
\centering
\caption{
This shows that performing multi-subspace updates in KL-Shampoo (ours, BFP16, QR, greedy) can improve accuracy at the cost of increased runtime.
We update the basis via multi-step procedures ($ K=1, 3, 5$) in subspaces. We consider different subspace sizes ($d_{\text{sub},i} = B d_i$). Cell colors encode the wall-clock runtime ratio relative to the full-basis QR baseline ($T=10$, $B=K=1$); see colorbar below.}
\label{tab:blocking}
\resizebox{\linewidth}{!}{%
\begin{tabular}{cccccc}
\toprule
$T$ &
$B$ & $K$ &
   \makecell{\textbf{nanoGPT} \\ \textbf{(123M)}}
  & \makecell{\textbf{llama3} \\ \textbf{(119M)}}
  & \makecell{\textbf{llama3} \\ \textbf{(313M)}}
  \\
\midrule
10 & $1$ & 1
  & \runtimeCell{$3.244$}{\baselineNanoGPTStep}{\baselineNanoGPTStep}
  & \runtimeCell{$3.345$}{\baselineLlamaSmallStep}{\baselineLlamaSmallStep}
  & \runtimeCell{$3.183$}{\baselineLlamaMid}{\baselineLlamaMid}
  \\
\midrule
 \multirow{3}{*}{10} & \multirow{3}{*}{$\nicefrac{1}{2}$} & 1
  & \runtimeCell{$3.247$}{649.95}{\baselineNanoGPTStep}
  & \runtimeCell{$3.344$}{0.4843}{\baselineLlamaSmallStep}
  & \runtimeCell{$3.187$}{1.0264}{\baselineLlamaMid}
  \\
  &  & 3
  & \runtimeCell{$3.245$}{724.22}{\baselineNanoGPTStep}
  & \runtimeCell{$3.344$}{0.5522}{\baselineLlamaSmallStep}
  & \runtimeCell{$3.187$}{1.1544}{\baselineLlamaMid}
  \\
  &   & 5
  & \runtimeCell{$3.244$}{804.49}{\baselineNanoGPTStep}
  & \runtimeCell{$3.345$}{0.6202}{\baselineLlamaSmallStep}
  & \runtimeCell{$3.184$}{1.2834}{\baselineLlamaMid}
  \\
\midrule
 \multirow{3}{*}{10} & \multirow{3}{*}{$\nicefrac{1}{4}$} & 1
  & \runtimeCell{$3.251$}{634.19}{\baselineNanoGPTStep}
  & \runtimeCell{$3.347$}{0.4714}{\baselineLlamaSmallStep}
  & \runtimeCell{$3.192$}{1.0025}{\baselineLlamaMid}
  \\
  &   & 3
  & \runtimeCell{$3.251$}{672.05}{\baselineNanoGPTStep}
  & \runtimeCell{$3.346$}{0.5082}{\baselineLlamaSmallStep}
  & \runtimeCell{$3.189$}{1.0574}{\baselineLlamaMid}
  \\
  & & 5
  & \runtimeCell{$3.248$}{711.85}{\baselineNanoGPTStep}
  & \runtimeCell{$3.345$}{0.5432}{\baselineLlamaSmallStep}
  & \runtimeCell{$3.184$}{1.1204}{\baselineLlamaMid}
  \\
\midrule
 \multirow{3}{*}{10} & \multirow{3}{*}{$\nicefrac{1}{6}$} & 1
  & \runtimeCell{$3.256$}{632.59}{\baselineNanoGPTStep}
  & \runtimeCell{$3.346$}{0.4693}{\baselineLlamaSmallStep}
  & \runtimeCell{$3.195$}{0.9925}{\baselineLlamaMid}
  \\
 &  & 3
  & \runtimeCell{$3.251$}{664.96}{\baselineNanoGPTStep}
  & \runtimeCell{$3.346$}{0.5062}{\baselineLlamaSmallStep}
  & \runtimeCell{$3.189$}{1.0284}{\baselineLlamaMid}
  \\
 &  & 5
  & \runtimeCell{$3.252$}{706.23}{\baselineNanoGPTStep}
  & \runtimeCell{$3.346$}{0.5342}{\baselineLlamaSmallStep}
  & \runtimeCell{$3.187$}{1.0684}{\baselineLlamaMid}
  \\
\bottomrule
\end{tabular}%
}

\vspace{0.5em}
\runtimeColorbarWorkshop
\end{minipage}\hfill
\begin{minipage}[t]{0.48\linewidth}
\centering
\caption{
This shows that updating the KL-Shampoo basis (ours, BFP16, QR, greedy) more frequently ($ T=2,3,4,5$) can be beneficial even when updating only a subset of the basis ($d_{\text{sub},i} = B d_i$).
A cheap subspace update can be useful when the basis is updated more frequently.
}
\label{tab:blocking2}
\resizebox{\linewidth}{!}{%
\begin{tabular}{cccccc}
\toprule
$T$ &
$B$ & $K$ &
   \makecell{\textbf{nanoGPT} \\ \textbf{(123M)}}
  & \makecell{\textbf{llama3} \\ \textbf{(119M)}}
  & \makecell{\textbf{llama3} \\ \textbf{(313M)}}
  \\
\midrule
 \multirow{3}{*}{2} & $\nicefrac{1}{3}$ & \multirow{3}{*}{1}
  & \runtimeCell{$3.247$}{723.19}{\baselineNanoGPTStep}
  & \runtimeCell{$3.344$}{0.5785}{\baselineLlamaSmallStep}
  & \runtimeCell{$3.183$}{1.199}{\baselineLlamaMid}
  \\
  & $\nicefrac{1}{4}$ &
  & \runtimeCell{$3.250$}{702.28}{\baselineNanoGPTStep}
  & \runtimeCell{$3.345$}{0.5545}{\baselineLlamaSmallStep}
  & \runtimeCell{$3.184$}{1.1545}{\baselineLlamaMid}
  \\
  & $\nicefrac{1}{5}$ &
  & \runtimeCell{$3.252$}{695.46}{\baselineNanoGPTStep}
  & \runtimeCell{$3.345$}{0.55}{\baselineLlamaSmallStep}
  & \runtimeCell{$3.184$}{1.129}{\baselineLlamaMid}
  \\
\midrule
 \multirow{3}{*}{3} & $\nicefrac{1}{3}$ &  \multirow{3}{*}{1}
  & \runtimeCell{$3.246$}{687.26}{\baselineNanoGPTStep}
  & \runtimeCell{$3.344$}{0.535}{\baselineLlamaSmallStep}
  & \runtimeCell{$3.184$}{1.119}{\baselineLlamaMid}
  \\
 & $\nicefrac{1}{4}$ &
  & \runtimeCell{$3.251$}{675.87}{\baselineNanoGPTStep}
  & \runtimeCell{$3.345$}{0.519}{\baselineLlamaSmallStep}
  & \runtimeCell{$3.184$}{1.089}{\baselineLlamaMid}
  \\
 & $\nicefrac{1}{5}$ &
  & \runtimeCell{$3.253$}{671.49}{\baselineNanoGPTStep}
  & \runtimeCell{$3.346$}{0.516}{\baselineLlamaSmallStep}
  & \runtimeCell{$3.186$}{1.072}{\baselineLlamaMid}
  \\
\midrule
  \multirow{3}{*}{4} & $\nicefrac{1}{3}$ & \multirow{3}{*}{1}
  & \runtimeCell{$3.247$}{668.98}{\baselineNanoGPTStep}
  & \runtimeCell{$3.345$}{0.51325}{\baselineLlamaSmallStep}
  & \runtimeCell{$3.184$}{1.0785}{\baselineLlamaMid}
  \\
   & $\nicefrac{1}{4}$ &
  & \runtimeCell{$3.252$}{660.85}{\baselineNanoGPTStep}
  & \runtimeCell{$3.346$}{0.50125}{\baselineLlamaSmallStep}
  & \runtimeCell{$3.188$}{1.05675}{\baselineLlamaMid}
  \\
  & $\nicefrac{1}{5}$ &
  & \runtimeCell{$3.250$}{659.23}{\baselineNanoGPTStep}
  & \runtimeCell{$3.346$}{0.499}{\baselineLlamaSmallStep}
  & \runtimeCell{$3.190$}{1.0435}{\baselineLlamaMid}
  \\
  \midrule
  \multirow{3}{*}{5} & $\nicefrac{1}{3}$ & \multirow{3}{*}{1}
  & \runtimeCell{$3.249$}{659.66}{\baselineNanoGPTStep}
  & \runtimeCell{$3.345$}{0.5002}{\baselineLlamaSmallStep}
  & \runtimeCell{$3.186$}{1.0544}{\baselineLlamaMid}
  \\
   & $\nicefrac{1}{4}$ &
  & \runtimeCell{$3.252$}{652.30}{\baselineNanoGPTStep}
  & \runtimeCell{$3.347$}{0.4906}{\baselineLlamaSmallStep}
  & \runtimeCell{$3.187$}{1.0372}{\baselineLlamaMid}
  \\
  & $\nicefrac{1}{5}$ &
  & \runtimeCell{$3.254$}{649.33}{\baselineNanoGPTStep}
  & \runtimeCell{$3.347$}{0.4888}{\baselineLlamaSmallStep}
  & \runtimeCell{$3.190$}{1.0264}{\baselineLlamaMid}
  \\
\bottomrule
\end{tabular}%
}
\end{minipage}
\end{table*}

\vspace{-0.1cm}
\paragraph{Experiment 4: Our reparametrization supports efficient subspace QR updates}
Block Jacobi methods often employ an inner loop and perform multi-step iterations (e.g., $K=1,3,5$) in subspaces. Given the similarity between these methods and ours, we investigate whether using the same loop is necessary for Shampoo-based methods.
\textbf{(I) With the inner loop:} We begin by fixing the decomposition frequency at $T=10$ for both full-basis and subspace updates. As shown in \cref{tab:blocking}, subspace updates can achieve performance comparable to that of full-basis updates when using the same loop, provided that the subspace is not too small. Using multi-step subspace updates can improve accuracy at the cost of increased runtime. By contrast, using a single step is inexpensive but inaccurate.
\textbf{(II) Without the inner loop:} 
Unlike the classical setting, $\mP_i$ is ever-changing rather than fixed. This motivates us to vary the decomposition frequency and use a single subspace update (i.e., $K=1$). This has a similar spirit to using QR decomposition as an approximation to eigendecomposition. Our experiments show (see \cref{tab:blocking2}) that a single subspace update can achieve decent performance relative to the full-basis method while reducing the total runtime. This finding highlights an accuracy-performance trade-off for improving Shampoo-based methods.

\vspace{-0.1cm}
\paragraph{Experiment 5: Using our reparametrization reduces the runtime of Shampoo-based methods}
In this set of experiments, we consider (pre-)training a larger model and demonstrate the benefits of using subspace updates. From the top-left panel of \cref{fig:time_cost}, we can see that subspace methods can outperform full-basis methods in runtime, at the cost of slightly degraded per-step performance.

\vspace{-0.2cm}
\section{Conclusion}

We introduce a reparametrization of the preconditioning matrices in Shampoo-based methods. This reparametrization supports half-precision storage while preserving performance. It also enables efficient updates of a subset of the orthogonal basis via QR decomposition in a subspace, thereby reducing the computational cost of QR. The approach applies broadly to Shampoo-based methods that employ QR decomposition. Our empirical results demonstrate its effectiveness and show that it improves the efficiency of these methods.

%

\bibliography{refs}
\bibliographystyle{plainnat}

\newpage
\appendix
\section{Make any QR Implementation Unique}
\label{app:qr_unique}
For a non-singular square matrix $\mA$, we can always make any QR implementation unique by requiring the upper-triangular matrix $\mR$ to have \emph{positive} diagonal entries.
This is possible because $\mR$ must have non-zero diagonal entries when $\mA$ is non-singular.
We then use the following procedure to make the decomposition unique
\begin{align*}
   \mathrm{qr}(\mA) = \underbrace{\mQ \mR}_{\text{non-unique QR}}  =
   \underbrace{\mQ \mD}_{=\bar{\mQ}} \underbrace{\mD^\top \mR}_{=\bar{\mR}}
\end{align*} where $\mD \mD^\top = \mI$ and  $\mD:=\mathrm{Diag}(\mathrm{sign}(\mathrm{diag}(\mR)))$ is constructed to be a diagonal sign matrix so that $\bar{\mR}:=\mD \mR$ has  positive diagonal entries. 
Note that $\bar{\mQ}$ is orthogonal and $\bar{\mR}$ is upper-triangular. Thus, we obtain a unique QR decomposition of $\mA$ via $\mA = \bar{\mQ} \bar{\mR}$.

\section{Additional Experimental Details}
\label{app:exp_details}

We conduct five sets of experiments on four language models—nanoGPT (123M) and Llama 3 (119M, 313M, and 598M)—using the FineWeb dataset from Hugging Face. We use the default train/test split in all experiments. As strong baselines for training matrix-valued weights in neural networks, we consider Muon \citep{liu2025muon} with the polar express backend \citep{amsel2026polar}, SOAP \citep{vyas2024soap}, and KL-Shampoo/SOAP \citep{lin2026understanding}. For vector-valued weights, such as those in normalization layers, we use AdamW.

In the nanoGPT experiments \citep{nanoGPT2024}, we use a constant learning rate with linear warm-up and cool-down, a batch size of 512, and a sequence length of 1024. In all Llama 3 experiments \citep{semenov2025benchmarking}, we use cosine learning-rate scheduling, a batch size of 768, and a sequence length of 512. For each method, we tune all available hyperparameters using random search with 120 runs. Our search follows a two-stage policy. In Stage 1, we explore a wider search range with 60 runs and narrow the range based on test loss. In Stage 2, we refine the search range and perform an additional 60 runs. For smaller models, namely nanoGPT (123M) and Llama 3 (119M), we train on four L40S GPUs. For larger models, namely Llama 3 (313M) and Llama 3 (598M), we train on two H100 GPUs. Due to the limited computational resources, we train each model for 10,000 iterations and report the performance of each method.

\section{Limitations and Future Work}
\label{app:future}

In this paper, we show that our method narrows the runtime performance gap between KL-Shampoo and Muon. Unfortunately, it still does not outperform Muon in runtime. This remains a limitation of the current submission.

However, we believe that our reparametrization can further reduce runtime by combining full-basis updates with subspace updates—for example, by using cheap subspace updates together with occasional expensive full-basis updates. Another promising direction is to use the adaptive decomposition frequency suggested by \citet{eschenhagen2025purifying}. This approach requires computing $\mP_i$ to determine the update frequency automatically. Thus, our reparametrization enables an efficient implementation of this idea.

Finally, we can adapt the parallel-subspace techniques studied in the Jacobi literature and perform QR updates on multiple smaller non-overlapping subspaces in parallel to further reduce runtime. Our reparametrization naturally supports these extensions. With these improvements, we believe KL-Shampoo can be made even faster and may further close the runtime gap in future work.

\end{document}